\documentclass[11pt]{article}

\usepackage[preprint]{acl}

\usepackage{times}
\usepackage{latexsym}
\usepackage{algorithm}
\usepackage{algpseudocode}
\usepackage[T1]{fontenc}
\usepackage[utf8]{inputenc}
\usepackage{microtype}
\usepackage{inconsolata}
\usepackage{graphicx}
\usepackage{xcolor}
\usepackage[normalem]{ulem} 
\usepackage{amsmath} 
\usepackage{amssymb} 
\usepackage{booktabs}
\usepackage{tcolorbox}
\tcbuselibrary{skins,breakable}

\definecolor{muhan}{HTML}{D97706}

\definecolor{kh}{HTML}{168aff}

\definecolor{hz}{HTML}{7E22CE}

\title{\textsc{Re-Search}: Dissecting LLM Ideation \\Through a Two-Player Questioning Game}

\author{
  Muhan Gao\textsuperscript{1} \quad
  Hangxiao Zhu\textsuperscript{1} \quad
  Xinqi Su\textsuperscript{1} \quad
  Yu Zhang\textsuperscript{1} \\
  \textbf{Daniel Khashabi}\textsuperscript{2,\dag} \quad
  \textbf{Kuan-Hao Huang}\textsuperscript{1,\dag} \\
  \textsuperscript{1}Texas A\&M University \quad
  \textsuperscript{2}Johns Hopkins University
}

\begin{document}
\maketitle
{\renewcommand{\thefootnote}{\dag}\footnotetext{Equal advising.}}

\begin{abstract}
Research ideation is iterative: researchers refine their ideas step by step based on feedback. Yet existing evaluations obscure this search process, making it difficult to understand how LLMs explore the space of research ideas and what limits their progress.
We introduce \textsc{Re-Search}, which adapts the format of Twenty Questions to research ideation: given only the research problem of a published paper, a model must rediscover the paper’s core idea by asking yes-or-no questions. Built from 200 ICML 2026 papers, \textsc{Re-Search} tracks how a model’s idea evolves as it asks questions and receives binary oracle feedback, isolating ideation from execution. Across six recent LLMs, nearly all progress happens within the first 30 of 100 questions, which largely decide how the game ends; after that, most models plateau. Through controlled interventions on the candidate questions and the hypothesis space, we find that the bottleneck is generation, not judgment: models recognize a good idea or a good question when one is offered, but often cannot produce it on their own. We further analyze the ideation traces and find that models fail by either over-exploiting one direction or over-exploring too many. These findings point to concrete design principles for autoresearch systems: since models can judge candidates they cannot produce, enriching the search with externally sourced ideas and questions may be more effective than improving any single model’s generation ability.
\end{abstract}


\section{Introduction}
\label{sec:intro}


\newcommand{\figgood}[1]{{\setlength{\fboxsep}{1pt}\colorbox{figgoodbg}{\textcolor{figgood}{\textbf{#1}}}}}
\newcommand{\figbad}[1]{{\setlength{\fboxsep}{1pt}\colorbox{figbadbg}{\textcolor{figbad}{\textbf{#1}}}}}
\definecolor{figgood}{HTML}{2E6A2A}
\definecolor{figbad}{HTML}{A6412A}
\definecolor{figgoodbg}{HTML}{E0EADD}
\definecolor{figbadbg}{HTML}{F7E2DC}
\definecolor{inqfill}{HTML}{0072B2}
\definecolor{ansfill}{HTML}{E69F00}
\definecolor{judfill}{HTML}{CC79A7}
\definecolor{inqcolor}{HTML}{0072B2}
\definecolor{anscolor}{HTML}{9A6B00}
\definecolor{judcolor}{HTML}{A16084}
\newcommand{\inq}[1]{\textcolor{inqcolor}{#1}}
\newcommand{\ans}[1]{\textcolor{anscolor}{#1}}
\newcommand{\jud}[1]{\textcolor{judcolor}{#1}}
\newcommand{\Inq}{{\color{inqcolor}M_{\mathrm{i}}}}
\newcommand{\Ans}{{\color{anscolor}M_{\mathrm{a}}}}
\newcommand{\Jud}[1][]{{\color{judcolor}M_{\mathrm{j}}^{#1}}}
\begin{figure}[!tp]
  \centering
  \includegraphics[width=\columnwidth]{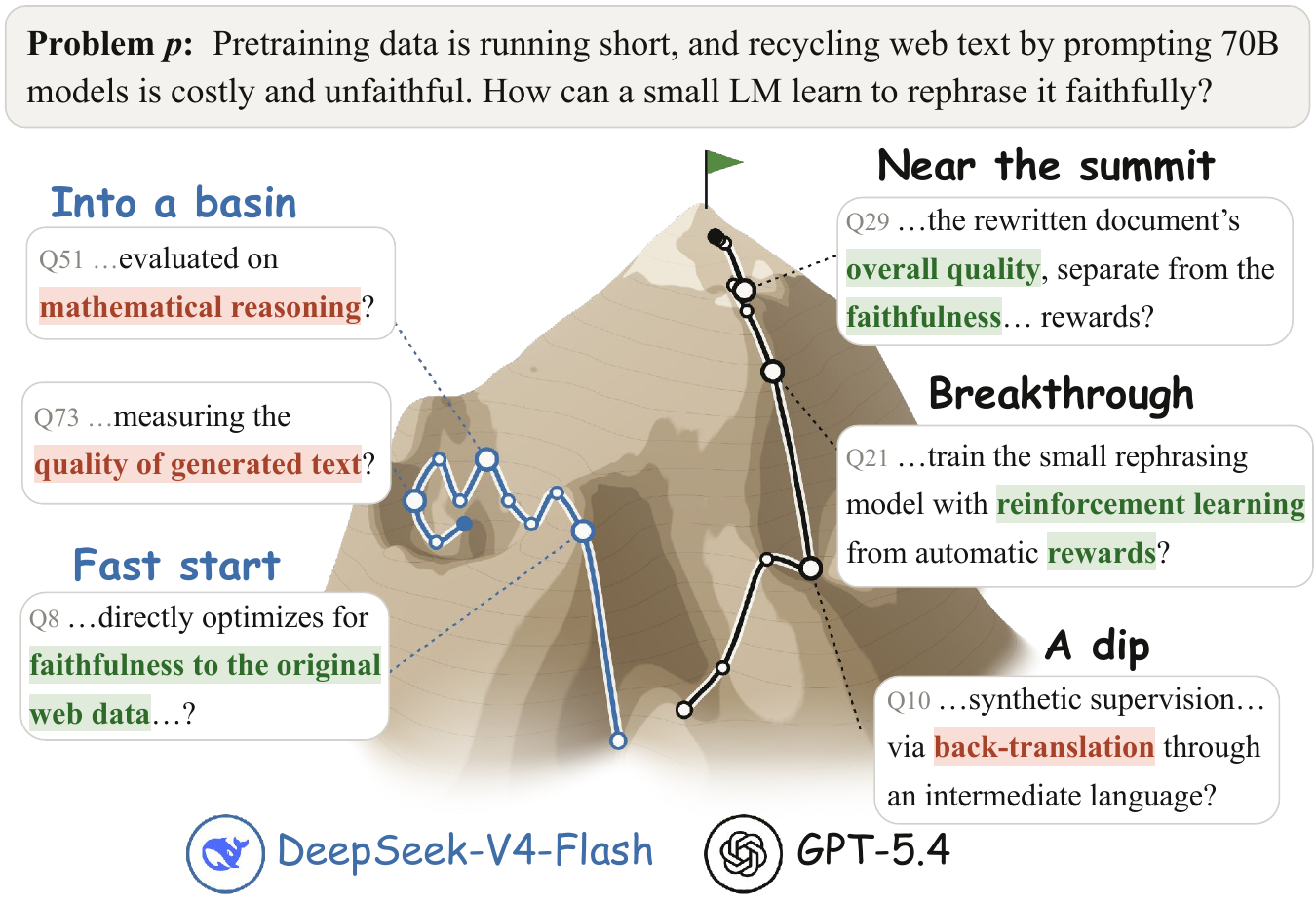}
    \caption{
    \texttt{DeepSeek-V4-Flash} and \texttt{GPT-5.4} play \textsc{Re-Search} on the same paper. Height reflects recovery of the paper’s idea. Green marks matching terms; red marks mismatches. While \texttt{DeepSeek-V4-Flash} ends up circling in a basin, \texttt{GPT-5.4} recovers from a dip and reaches the summit. 
    }
  \label{fig:climb}
\end{figure}

Research ideation is not a one-shot proposal but an iterative search \textit{interleaved} with every stage of a project: researchers form hypotheses, test them, and refine them based on feedback. This process can be viewed as \textit{adaptive information seeking} over a space of plausible yet unverified ideas~\citep{klahr1988dual}. 
However, existing benchmarks provide limited visibility into this iterative search. Ideation benchmarks evaluate one-shot proposals using LLM judges~\citep{guo2025ideabench,chen2025mlrbench}, overlooking how ideas evolve through feedback. End-to-end benchmarks involve iterative search but entangle it with coding, debugging, and experimentation~\citep{wijk2025rebench,wu2026innovatorbench,rank2026posttrainbench,lupidi2026airsbench,garikaparthi2026researchgym,chan2025mlebench}, making it difficult to distinguish failures of ideation from failures of execution.

\begin{figure}[t]
  \centering
  \includegraphics[width=\columnwidth]{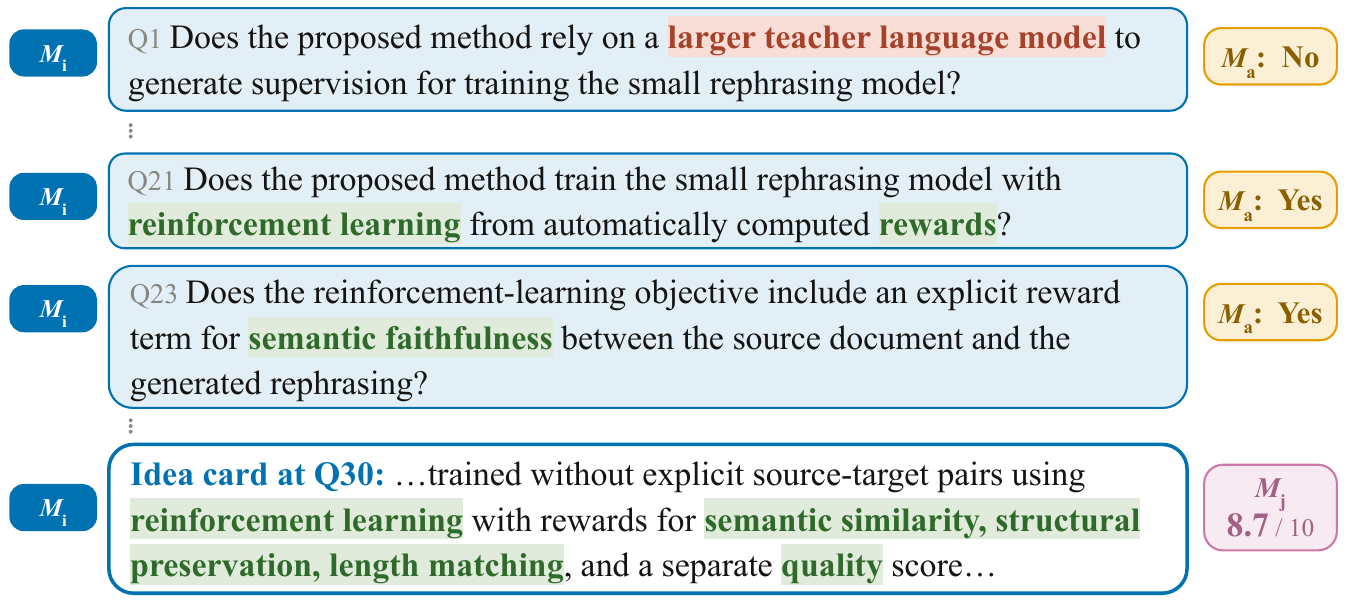}
  \caption{Part of the \texttt{GPT-5.4} game in Figure~\ref{fig:climb}: given the research problem $p$, the \inq{inquisitor} $\Inq$ asks yes-or-no questions, the \ans{answerer} $\Ans$ replies based on the full paper, and the \jud{judges} $\Jud$ score the idea card.}
  \label{fig:qa}
\end{figure}

In this work, we aim to isolate the information-seeking process from the rest of autoresearch. To this end, we introduce \textsc{Re-Search}, a two-player questioning game that makes every step of this process observable and measurable (Fig~\ref{fig:climb}). The setting is a scientific variant of Twenty Questions~\citep{mosher1966asking,siegler1977twenty}, a classic paradigm in cognitive science for studying human reasoning and problem solving~\citep{coenen2019asking}. Given only the research problem of a published paper, an inquisitor model must rediscover the paper’s idea by asking up to 100 yes-or-no questions, while an answerer model reads the full paper and replies to the questions. Every 10 questions, the inquisitor model writes down its current best guess of the idea, which a panel of judges scores against the paper’s idea (\S\ref{sec:research}, see Fig~\ref{fig:qa} as an example). Success on \textsc{Re-Search} requires three capabilities.
The model must (1) ask \emph{informative questions}, which means both proposing good candidate questions and judging which of them to ask,
(2) maintain an \emph{adequate hypothesis space} that includes the correct idea,
and (3) \emph{hill-climb} along promising directions while abandoning dead ends.



We evaluate six recent LLMs on 200 ICML 2026 papers. GPT-5.4 leads by a wide margin, followed by Gemma-4-31B and Claude Sonnet 4.5; the remaining three models score about two thirds as high as GPT-5.4. Yet the models share a common pattern: most progress happens within the first 30 questions, after which gains slow sharply for all but GPT-5.4, and a model’s score at question 30 largely predicts its final score. So why do models stall with 70 questions left, and which capabilities are missing?

We first rule out a simple knowledge deficit: on papers where both a strong and a weak model can list the core idea beforehand, the strong model still finishes far ahead. The first clue comes from \textit{question quality}. When a weak model can choose from candidate questions proposed by several models, it scores far above its own solo performance. The same pattern holds for the \textit{hypothesis space}. We give models a survey of candidate directions that includes the paper's core idea; without it, many models never reach the idea, but with it, they zero in on it within a few questions---and the weaker the model, the larger the gain.

In both cases, models \emph{recognize} a good idea or a good question when one is offered, but often cannot \emph{generate} it on their own. The ideation traces show how \textit{hill-climbing} fails: weaker models either over-exploit one direction or over-explore too many, stuck into an unproductive pattern early.


Our contributions are: (1) \textsc{Re-Search}, a two-player questioning game that isolates research ideation from execution and makes every step of ideation observable and measurable; (2) the finding that ideation is decided early: most progress happens within the first 30 questions, and weaker models lock into either over-exploiting one direction or over-exploring too many; and (3) a diagnosis that the bottleneck is generation, not judgment, pointing to external candidates as a direct lever for improving autoresearch systems.

\section{Related Work}
\label{sec:related}
\paragraph{AI scientists and their evaluation.}
LLM-based systems now attempt to automate research from idea to paper~\citep{lu2024aiscientist,lu2026aiscientist,schmidgall2025agentlab,gottweis2026coscientist}.
Most are evaluated end to end:
benchmarks run an agent on a research project for hours and score the outcome~\citep{chan2025mlebench,wijk2025rebench,zhang2025mlrcbench,wu2026innovatorbench,rank2026posttrainbench,lupidi2026airsbench,garikaparthi2026researchgym},
or grade intermediate steps against expert rubrics~\citep{starace2025paperbench,kon2026expbench,bragg2026astabench,jansen2024discoveryworld},
and open-world environments let agents pursue long projects on their own~\citep{chung2025station}.
Some of them ask agents to rediscover the findings of a hidden paper, from raw data~\citep{xu2026researchclawbench} or by running experiments~\citep{wang2026firebench}.
In all of these, the search for the idea is mixed with coding, debugging, and experiments,
and can only be reconstructed from logs afterwards~\citep{li2026beyond}.
\textsc{Re-Search} isolates this search, so that each question and its effect on the idea can be observed directly.

\paragraph{Research ideation.}
A separate line of work asks whether LLMs can propose research ideas.
\citet{si2025ideas} find in a large human study that LLM ideas are judged more novel but slightly less feasible than those of experts,
and benchmarks such as IdeaBench~\citep{guo2025ideabench} and the ideation stage of MLR-Bench~\citep{chen2025mlrbench} rate generated ideas with LLM judges.
Other systems refine ideas over several rounds,
through literature-grounded review~\citep{baek2025researchagent}, planned retrieval of external knowledge~\citep{hu2025nova},
or multi-agent and quality-diversity search~\citep{chen2026combinatorial,gumma2026ideagent}.
These evaluations either treat ideation as a single step or guide the search with LLM feedback,
and none of them has a correct answer to compare against.
Like us, \citet{yang2025moosechem} and \citet{liu2026researchbench} use recent papers as ground truth,
testing whether LLMs can rediscover their hypotheses by retrieving inspirations and composing a hypothesis from them.
The model receives its evidence up front, whereas in \textsc{Re-Search} it must find the evidence itself by asking. 

Concurrently, \citet{wen2026priced} introduce Priced Guidance to quantify the cost of steering a model toward a paper's core idea. The evaluated model proposes multiple-choice questions with reported answer probabilities. A guide with access to the target then selects an answer, priced at $-\log_2 p_{\mathrm{selected}}$ bits. Because this cost is computed from verbalized probabilities, the resulting score reflects probability-reporting behavior as well as candidate generation. Their analysis also highlights the cost of acquiring useful candidate concepts, consistent with our finding that weaker models can use supplied candidates they struggle to propose independently. In our work, the focus is on \emph{how} ideation unfolds through interaction and \emph{why} progress stalls. Following the cognitive view of discovery as hypothesis search guided by experimental feedback~\citep{langley1987discovery}, \textsc{Re-Search} tracks idea recovery through model-generated questions and paper-grounded feedback. Controlled interventions on candidate questions and hypotheses then examine how candidate availability and selection affect recovery, without requiring models to report answer probabilities.

\paragraph{Information seeking and question asking.}
Twenty Questions and related games are standard tools for studying how people ask questions:
children move from testing single hypotheses to asking questions that narrow the space~\citep{mosher1966asking,ruggeri2015children,ruggeri2016sources},
and people adapt their questions to the structure of the hypothesis space~\citep{nelson2014children,cohen2016searching,coenen2019asking}.
Recent work uses the same games to evaluate and improve LLMs~\citep{bertolazzi2023chatgpt,zhang2024probing,mazzaccara2024learning},
often by choosing questions with high expected information gain or value of information~\citep{hu2024uncertainty,choudhury2026bedllm,grand2026shoot,dong2026value};
MediQ~\citep{li2024mediq} brings the same setting to clinical diagnosis.
In these tasks, the target is an object, a diagnosis, or a board configuration from a small, known set of candidates.
In \textsc{Re-Search}, the target is a research idea, and the model must generate the candidates itself.
More broadly, LLMs in multi-turn and long-horizon settings tend to commit early and rarely recover from early mistakes~\citep{laban2026lost,sinha2026illusion},
although deep-research agents can also switch direction mid-trajectory~\citep{li2026laststep};
we find that in research ideation, early questions largely decide the outcome (\S\ref{sec:prefix}).



\section{The \textsc{Re-Search} Game}
\label{sec:research}

We define the game in \S\ref{sec:task}, describe how we build it from published papers in \S\ref{sec:data},
and explain how we score what the model recovers in \S\ref{sec:eval}.

\subsection{Task Description}
\label{sec:task}


In Twenty Questions, one player holds a hidden target (e.g., a specific person), and the other must identify it by iteratively asking yes-or-no questions.
\textsc{Re-Search} adapts this format to research ideation. Each game is played on a pair $(p, s)$ built from a published paper in ICML 2026, where $p$ is the research problem and $s$ is the solution that the paper proposes for it.

Three roles are involved:
(1) The \emph{\inq{inquisitor}} $\Inq$ sees only $p$.
At each turn $t$, it asks a yes-or-no question $q_t$ based on the problem and the history of earlier questions and answers,
$q_t = \Inq(p, h_{t-1})$, where $h_{t-1} = (q_1, a_1, \dots, q_{t-1}, a_{t-1})$ and $h_0$ is empty.
(2) The \emph{\ans{answerer}} $\Ans$ reads the full paper $d$ and replies to each question on its own,
$a_t = \Ans(d, q_t) \in \{\textsc{Yes}, \textsc{No}, \textsc{Unknown}\}$.
At fixed points $t \in \{0, 10, \dots, T\}$, the inquisitor also writes an idea card $c_t = \Inq(p, h_t)$,
a short, structured statement of its current best guess of $s$.

(3) After the game, a panel of three LLM \jud{judges} $\Jud[(1)], \Jud[(2)], \Jud[(3)]$ scores all saved idea cards $\{c_0, c_{10}, \dots, c_T\}$ against $s$.





\paragraph{Information revealed by questioning.}
Let $S$ be the hidden solution, $P$ the research problem, and $H_T$ the question--answer history. Suppose answers belong to a finite alphabet $\mathcal A$, and each question is conditionally independent of $S$ given $P$ and the preceding history. Then
\[
I(S;H_T\mid P) \leq T\log_2|\mathcal A|.
\]
Binary feedback yields at most $T\log_2 2=T$ bits; our three-answer alphabet yields at most $T\log_2 3$ bits. This bounds information gained through questioning, conditional on the problem, rather than information supplied before the game. Additional questions need not supply additional information, and this bound does not guarantee a particular recovery score. See \S\ref{app:information} for the proof.


\begin{algorithm}[t]
  \caption{The \textsc{Re-Search} game for one paper.}
  \label{alg:game}
  \small
  \begin{algorithmic}[1]
    \Require research problem $p$, paper $d$, hidden solution $s$, question budget $T$
    \State $h_0 \gets ()$
    \State $c_0 \gets \Inq(p, h_0)$ \Comment{idea card before any question}
    \For{$t = 1, \ldots, T$}
      \State $q_t \gets \Inq(p, h_{t-1})$
      \State $a_t \gets \Ans(d, q_t) \in \{\textsc{Yes}, \textsc{No}, \textsc{Unknown}\}$ \Comment{stateless}
      \State $h_t \gets (h_{t-1}, q_t, a_t)$
      \If{$t \in \{10, 20, \ldots, T\}$}
        \State $c_t \gets \Inq(p, h_t)$
      \EndIf
    \EndFor
    \State $\Jud[(k)]$ score every saved $c_t$ against $s$ \Comment{not fed back}
  \end{algorithmic}
\end{algorithm}

\subsection{Data Construction} 
\label{sec:data}
We build \textsc{Re-Search} from 200 papers accepted at ICML 2026, one of the most recent top-tier conferences. The papers are selected based on three criteria: (1) the main contribution is a method, rather than a benchmark, a dataset, or an analysis, so that each paper has a concrete solution to recover; (2) its review score is as high as possible, indicating a reliable solution; and
(3) the paper became public as late as possible. All selected papers first appeared, on arXiv or the conference, after 31 August 2025,
later than every reported knowledge cutoff (\S\ref{app:cutoff}), minimizing the risk of knowledge contamination~\citep{cheng2024dated,roberts2024cutoff}.

For each paper, we take its full text $d$ and use an LLM (GLM-5.2) to extract the pair $(p, s)$ in a structured form,
which has been shown to be reliable in prior work~\citep{gao2025panoramas,liu2026researchbench}.
The research problem $p$ has four fields (the problem domain, the research goal, the main challenge, and the gap in prior work),
and the solution $s$ has four fields (the main contribution, the approach, the technical components, and the novelty). Human annotators (three Computer Science PhD students) checked the extracted pairs $(p, s)$ for 100 papers. By majority vote, 93\% are correct, and the errors mostly omit a component that distinguishes the paper's method.



\subsection{Evaluation}
\label{sec:eval}
We score each idea card $c_t$ against the hidden solution $s$ with a panel of three LLM judges
(Gemma-4-31B, Qwen3.6, and GPT-OSS-120B).
Each judge sees the research problem $p$, hidden solution $s$, and idea card $c_t$, and rates on a 1--10 scale how much of $s$ the card recovers.
The recovery score is the mean of the three ratings, $r_t = \frac{1}{3}\sum_{k=1}^{3} \Jud[(k)](p, s, c_t)$.
The judges agree closely: over all the cards in our experiment,
their pairwise Spearman correlations range from 0.85 to 0.88.
Since three of the judges are also inquisitors, we check for self-preference: leaving out the judge from an inquisitor's own model family changes the gain from naming the idea by at most 0.08 points (\S\ref{app:selfeval}).

\textbf{Human evaluation.} We use human annotators to check the reliability of the LLM judges, by comparing 498 pairs of idea cards;
for each pair, they choose the card closer to the hidden solution, or mark a tie.
Their labels agree well, with a Fleiss' $\kappa$ of 0.83.
In 91.2\% of pairs, the LLM judges agree with the annotator majority on which card is closer, or that the two are tied.
Most of the remaining pairs are ones that the annotators mark as a tie but the judges do not;
when both sides prefer one card, they choose the same card in 97.8\% of pairs.


\begin{figure}[!t]
    \centering
    \includegraphics[width=\linewidth, trim={0 5 0 0}, clip]{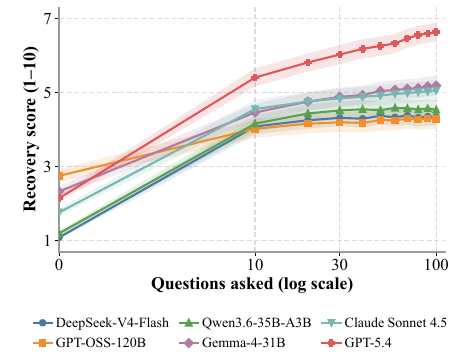}
    \caption{Recovery score of the games over 100 rounds, averaged over 200 papers and three games per paper. The patterns of models are different: three models gain little after the first 30 questions, Gemma and Sonnet slowly climb, while GPT-5.4 has consistent improvement. (\S\ref{sec:stall})}
    \label{fig:b0}
\end{figure}

\section{Ideation Dynamics in \textsc{Re-Search}}
\label{sec:exp-result}

\S\ref{sec:exp-result} to \S\ref{sec:tree} build on each other. In this section, we describe what happens in a game; \S\ref{sec:why} asks why most models plateau early, with controlled interventions on the candidate questions (\textit{informative questions}) and the ideas it considers (\textit{hypothesis space}); and \S\ref{sec:tree} examines how models follow promising directions and abandon dead ends (\textit{hill-climbing}), through the shape of the trace. We begin with the model choices, then show how ideation develops over a game (\S\ref{sec:stall}) and that the early questions largely decide its outcome (\S\ref{sec:prefix}). 
\paragraph{Model selection.}
We evaluate both open-weight and proprietary LLMs of different strengths:
Gemma-4-31B, GPT-OSS-120B, DeepSeek-V4-Flash-Preview, Qwen3.6-35B-A3B,
Claude Sonnet 4.5, and GPT-5.4. We do not include newer models because of the risk of knowledge contamination.
In all games, Gemma-4-31B serves as the answerer,
each inquisitor asks up to $T = 100$ questions, and we play three games per paper. Among three candidate answerers (GPT-OSS-120B, Qwen3.6-35B-A3B, and Gemma-4-31B),
Gemma agrees most often with human labels on 795 annotated answers
(94.0\%, against 86.2\% for GPT-OSS-120B and 78.4\% for Qwen3.6-35B-A3B; see \S~\ref{app:answerer} for details),
and only 1.1\% flip \textsc{Yes} and \textsc{No},
far below the 10\% answer noise assumed in~\citet{grand2026shoot}. Therefore we use Gemma-4-31B as the answerer model.

\begin{figure*}[t]
  \centering
  \includegraphics[width=\textwidth]{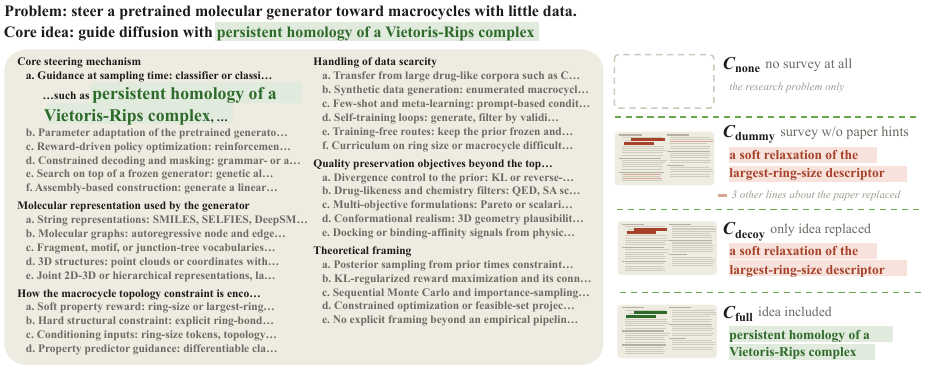}
  \caption{An example survey under $C_{\text{full}}$.
  \textbf{Left:} the $C_{\text{full}}$ survey given before the game,
  which names the core idea as one option, with no extra hint.
  \textbf{Right:} how the four context conditions differ.
  $C_{\text{decoy}}$ replaces only that phrase with a plausible decoy,
  $C_{\text{dummy}}$ also replaces the other lines that point to the paper, and $C_{\text{none}}$ gives no survey.
  Because $C_{\text{full}}$ and $C_{\text{decoy}}$ differ in a single phrase,
  any difference between them only comes from whether the core idea is in the survey. (\S\ref{sec:menu})}
  \label{fig:menu-example}
\end{figure*}

\subsection{Most Models Stall Early}
\label{sec:stall}

Figure~\ref{fig:b0} shows how recovery develops over a game. Before the first question, every model's idea card $c_0$ scores very low, confirming that the research problem alone does not reveal the solution.
The game then unfolds in three stages.

\emph{Questions 1--10} bring the largest gain for every model, with scores rising by $+1.3$ to $+3.2$.
\emph{Questions 11--30} add much less, between $+0.2$ and $+0.8$.
The most interesting part happens in \emph{Questions 31--100}, where the models are separated into different patterns.
Three models stall, gaining almost nothing over the remaining 70 questions; two keep improving slowly; and only the strongest model keeps climbing steadily.

\subsection{Early Questions Decide the Outcome}
\label{sec:prefix}

\begin{table}[t]
  \centering
  \small
  \setlength{\tabcolsep}{3pt}
  \begin{tabular}{@{}lcccc@{}}
    \toprule
    Model & $k$ & Gap at $k$ & Gap at 100 & Ahead (\%) \\
    \midrule
                      & 10 & 1.59 & 1.16 & 63 \\
    Gemma-4-31B       & 20 & 1.53 & 1.64 & 70 \\
                      & 30 & 2.28 & 2.30 & 78 \\
    \midrule
                      & 10 & 1.03 & 1.00 & 64 \\
    Qwen3.6-35B-A3B   & 20 & 1.21 & 1.41 & 69 \\
                      & 30 & 1.86 & 1.70 & 73 \\
    \midrule
                      & 10 & 1.70 & 1.85 & 68 \\
    GPT-OSS-120B      & 20 & 2.32 & 2.45 & 78 \\
                      & 30 & 2.56 & 2.54 & 77 \\
    \midrule
                      & 10 & 1.48 & 1.37 & 66 \\
    DeepSeek-V4-Flash & 20 & 2.10 & 1.72 & 71 \\
                      & 30 & 2.14 & 2.20 & 79 \\
    \bottomrule
  \end{tabular}
  \caption{Prefix-fork experiment: we truncate the best and worst game at round $k$ and continue each three times. \emph{Gap at $k$} means score difference between the best and worst prefix at round k. \emph{Ahead}: share of papers where the best prefix still leads at question 100. The gap largely persists across models. (\S\ref{sec:prefix})}
  \label{tab:prefix}
\end{table}
A natural implication of \S\ref{sec:stall} is that, if most models stall after about 30 questions,
the early questions should largely decide how a game ends.
This echoes findings in long-horizon LLM generation,
where early mistakes compound and are rarely repaired~\citep{zhang2024snowball,laban2026lost,sinha2026illusion}. We test this with a prefix-fork experiment~\citep{bigelow2025forking}. Each model plays three games per paper. First, for each paper, we pick the games with the highest and the lowest final score, and truncate both at round $k$ ($k \in \{10, 20, 30\}$). Then, the same model continues each prefix to question 100 three more times, yielding three continuations from the best start and three from the worst. If the opening matters, the best-start continuations should stay ahead.

As shown in Table~\ref{tab:prefix}, the score at question $k$ predicts the score at question 100 well.
Across games, the correlation is $r = 0.76$--$0.87$ at $k = 10$ and rises to $r = 0.87$--$0.92$ at $k = 30$.

\section{Diagnosing the Bottleneck}
\label{sec:why}
As \S\ref{sec:exp-result} shows, most models plateau early, while GPT-5.4 keeps climbing. What explains this gap? We first rule out knowledge as the main cause (\S\ref{sec:knowledge}), then show that supplying external candidates, either as questions from other models (\S\ref{sec:pool}) or as a survey of the hypothesis space (\S\ref{sec:menu}), closes much of the gap.

\subsection{Knowledge Explains Little of the Gap}
\label{sec:knowledge}
A strong model may simply have seen more scientific text,
know more candidate ideas before the game, and turn this knowledge into an advantage in the game. To test this, we use a closed-book probe that isolates the effect of knowledge.
We collect a subset of 100 papers, and focus on two models, a strong one (GPT-5.4) and a weak one (GPT-OSS-120B). Each model is provided with the research problem and asked to list ten candidate solutions,
three times, for 30 candidates in total; two LLM annotators then judge whether the list contains the paper's core idea. 

Knowledge does not explain the gap between models. On this 100 papers subset, GPT-5.4 scores 2.54 points higher than GPT-OSS-120B; on the 24 papers where both models list the core idea, it still scores 2.28 higher, and  knowledge closes barely 10\% of the gap. Therefore, knowing the idea is not enough.

\subsection{Weak Models Can Pick Informative Questions}
\label{sec:pool}

\S\ref{sec:knowledge} suggests that the gap lies in how models ask. Asking an informative question involves two steps: \textbf{proposing} candidate questions and \textbf{choosing} which one to ask. To separate the two, we let a weak model choose each question from candidates proposed by several models, rather than asking only its own. If it improves, the bottleneck lies in proposing rather than judging. We test this using GPT-5.4, Qwen3.6-35B-A3B, DeepSeek-V4-Flash, and GPT-OSS-120B. Each round, all four models independently propose a question from the same history, and one model picks among them.

\begin{figure}[t]
  \centering
  \includegraphics[width=\columnwidth]{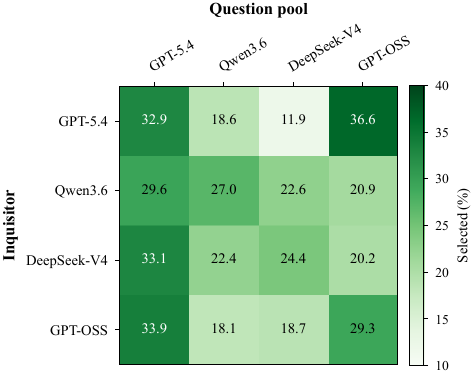}
  \caption{Share of questions (\%) each inquisitor (row) chooses from each proposer (column)
    in the four-model pool on 200 papers. All models favor questions from GPT-5.4 and themselves. While the proportions differ across inquisitors, the final scores are close. All four score between 5.19 and 5.34, with no significant difference.
     (\S\ref{sec:pool})}
  \label{fig:selector}
\end{figure}

After 30 rounds with the shared pool, all four models score between 5.19 and 5.34, regardless of which model picks the questions. The weaker models gain over 1 points compared to their solo baseline. Surprisingly, the gain of weak models does not come from blindly copying GPT-5.4's questions. For example, GPT-OSS-120B picks its own proposal in 30\% of rounds and GPT-5.4's in 34\% (Figure~\ref{fig:selector}).

\begin{figure}[!t]
  \centering
  \includegraphics[width=\columnwidth]{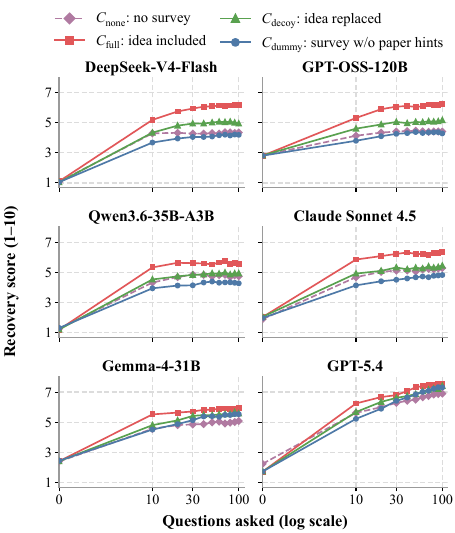}
  \caption{Recovery curves under the four context conditions of Figure~\ref{fig:menu-example} (200 ICML 2026 papers, three games per paper). Naming the idea ($C_{\text{full}}$) lifts the curves of most models clearly above $C_{\text{decoy}}$ and $C_{\text{dummy}}$; for GPT-5.4, all four conditions end close together. (\S\ref{sec:menu})
  }
  \label{fig:results}
\end{figure}


Which of the two steps drives the gain? We vary each while holding the other fixed. When GPT-OSS-120B proposes all four candidates itself (best-of-4 sampling, akin to self-consistency~\citep{wang2023selfconsistency}), its score rises only from 4.16 to 4.57, well below the 5.20 where candidates are from four models, so the candidates matter. When we keep the four-model pool but pick at random, the score drops to 4.43, not significantly above playing alone, so selection matters too (Table~\ref{tab:selector}). Both steps are necessary, but they are not equally scarce. When we swap the selector across models, the final scores barely change (Table~\ref{tab:selector}); the gap between models therefore comes mainly from the candidates they propose. A weak model’s selection is good enough; what it lacks are \textit{informative questions} worth selecting.

\begin{table}[htbp]
  \centering
  \small
  \setlength{\tabcolsep}{3pt}
  \newcommand{\ci}[2]{\,{\scriptsize[$#1$, $#2$]}}
  \begin{tabular*}{\columnwidth}{@{\extracolsep{\fill}}lrl@{}}
    \toprule
    Inquisitor & Score & Gain over alone \\
    \midrule
    GPT-OSS-120B      & 5.20 & $+1.04$\ci{+0.74}{+1.34} \\
    Qwen3.6-35B-A3B   & 5.34 & $+1.09$\ci{+0.87}{+1.39}  \\
    DeepSeek-V4-Flash & 5.24 & $+1.07$\ci{+0.79}{+1.36} \\
    GPT-5.4           & 5.19 & $-0.59$\ci{-0.90}{-0.28} \\
    \midrule
    Random pick       & 4.43 & --- \\
    \bottomrule
  \end{tabular*}
  \caption{Score after 30 questions for each inquisitor on the same four-model pool on 200 papers.
    The last column gives the difference from each model's solo baseline, with 95\% confidence intervals. (\S\ref{sec:pool})}
  \label{tab:selector}
\end{table}

\subsection{A Survey Enriches the Hypothesis Space}
\label{sec:menu}

\begin{figure}[t]
  \centering
  \includegraphics[width=\columnwidth]{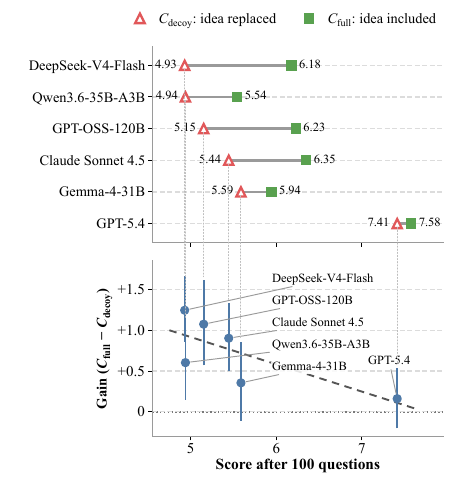}
  \caption{\textbf{Top:} score after 100 questions under $C_{\text{decoy}}$ and $C_{\text{full}}$ for each inquisitor.
    \textbf{Bottom:} gain in that score from naming the idea ($C_{\text{full}} - C_{\text{decoy}}$, with 95\% confidence intervals),
    against the score under $C_{\text{decoy}}$; the dashed line is a least-squares fit
    (Pearson $r = -0.74$ across the six inquisitors).
    Naming the idea helps most the models that score lowest without it. (\S\ref{sec:menu})}
  \label{fig:recognize}
\end{figure}

In real research, ideas do not start from a purely blank page,
as researchers usually begin with a \textit{landscape} in mind:
a rough sense of which approaches could solve the problem.
We give the inquisitor a similar starting point with a \emph{survey} (Figure~\ref{fig:menu-example}), an overview of the hypothesis space for the research problem. Each survey has five to seven dimensions along which solutions to the problem can differ,
and lists four to six candidate approaches in each dimension, named at the level of known families of techniques. The survey is placed right after the problem when the model asks questions, with no instruction on how to use it, and is not shown when the model writes its idea card
(\S~\ref{app:map}). More importantly, the inquisitor is not told whether the solution is in the survey.

Figure~\ref{fig:results} shows the four context conditions.
Naming the core idea ($C_{\text{full}}$) lifts the curves of most models clearly above $C_{\text{decoy}}$ and $C_{\text{dummy}}$.
When the name is replaced by a decoy ($C_{\text{decoy}}$), the survey still helps through the other listed components,
but the curves slightly fall back towards $C_{\text{none}}$;
removing those components as well ($C_{\text{dummy}}$) brings them closer still.

A survey enlarges the model’s hypothesis space from outside: under $C_{\text{full}}$, the correct idea is already among the candidates. Naming the core idea in the survey makes every model reach the core idea during the game more often, and helps weaker models most: the final score improves by up to 1.25 points, with the largest gains for the models that score lowest without it ($r = -0.74$, Figure~\ref{fig:recognize}). This is not because models simply go through the survey item by item. A survey lists 20 to 42 candidates, yet under $C_{\text{full}}$ models ask about the named idea within a median of 3 to 6 questions, and far more often than they ask about a decoy in the same line under $C_{\text{decoy}}$ (67--89\% against 29--52\% of games; \S~\ref{app:map}). Even under $C_{\text{decoy}}$, models ask about that line more often than about nearby lines, so they can find the right family of methods; what the name adds is the specific idea. Once the correct idea is in the hypothesis space, weak models recognize it; what they lack is the ability to put it there themselves.





\section{Turning Ideation Traces into Trees}
\label{sec:tree}

\S\ref{sec:why} shows why models differ, but not how ideation unfolds question by question.
A game is recorded as a flat sequence of questions, but the questions are not independent:
most of them follow up on an earlier question, while the rest open new directions.
Research proceeds in the same way: a researcher opens a few lines of inquiry, runs follow-up experiments on the promising ones, and drops those that stop paying off.
We therefore turn each game into a tree that makes this structure explicit.
Each top-level branch is a line of inquiry, and its depth shows how far the model follows it; together they reflect the model’s \textit{hypothesis space} and its \textit{hill-climbing}.

As using LLMs to turn reasoning traces into trees or graphs has become a common way to analyze how LLMs reason~\citep{jiang2025lcot2tree,zeng2025rejump,lee2025reasoningflow},
we follow this line of work and use GPT-6 Astra to construct the trees.
For every question, the model chooses a parent:
one of the earlier questions, or the research problem if the question opens a new direction.

\begin{figure}[t]
  \centering
  \includegraphics[width=\columnwidth]{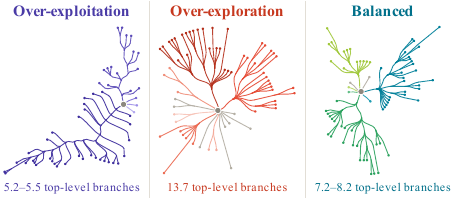}
  \caption{Three patterns of traces in games.
    The grey dot is the research problem,
    and each color is one top-level branch.
    Below each tree: mean top-level branches per game on 200 papers, as a range over the models with that pattern (per-model means in Figure~\ref{fig:app-tree-branches}). See more examples in \S\ref{app:trees}.
    }
  \label{fig:tree-types}
\end{figure}

\subsection{The Shape of the Search}
\label{sec:shape}
We build 1,200 trees from six models on the 200 papers. Although the three weaker models end with similar scores, their trees reveal opposite failure modes (Figure~\ref{fig:tree-types}). The number of top-level branches $B_{\mathrm{top}} = \sum_{t=1}^{T} \mathbf{1}[\operatorname{parent}(q_t)=p]$, i.e., questions asked directly under the research problem, cleanly separates three patterns (Figure~\ref{fig:tree-branches}).

\textbf{Over-exploitation.} DeepSeek-V4-Flash and Qwen3.6 open only 5.2--5.5 branches per game. Once they confirm a direction, they enumerate variants within it rather than questioning the direction itself. For example, after learning a paper uses weight statistics for quantization scales, DeepSeek-V4-Flash spends dozens of questions guessing which statistic, never asking whether the approach is something else entirely. Across all games, 23\% of these two models’ questions are such enumerations.

\textbf{Over-exploration.} GPT-OSS-120B opens 13.7 branches, jumping to new methods regardless of feedback. It is the only model that switches branch more often after a \textsc{Yes} than after a \textsc{No}.

\textbf{Balanced.} The three stronger models open 7.2--8.2 branches, following confirmed directions and stepping back when a line stops paying off.

While the two failure modes are opposite, they both explain why most models plateau after 30 questions (\S\ref{sec:stall}). The stronger models balance the two, which lets them keep climbing.

\subsection{Search Breadth and Final Recovery}
\begin{figure*}[t]
  \centering
  \begin{minipage}[b]{0.49\textwidth}\centering
    \includegraphics[width=\linewidth]{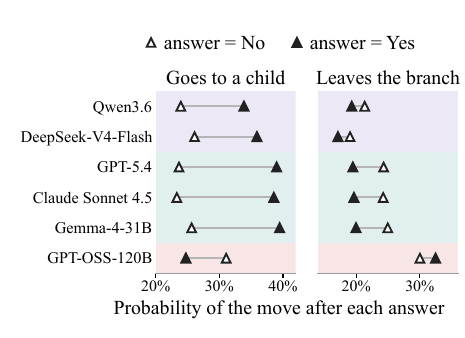}
  \end{minipage}\hfill
  \begin{minipage}[b]{0.49\textwidth}\centering
    \includegraphics[width=\linewidth]{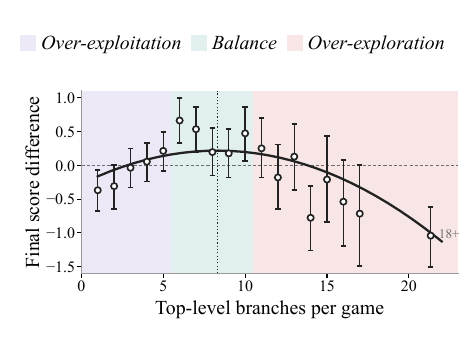}
  \end{minipage}
  \caption{Search breadth and final recovery across 1,200 games.   Shading separates the three patterns in \S\ref{sec:shape}.
  \textbf{Left.} After the inquisitor receives a Yes ($\blacktriangle$) or a No ($\vartriangle$), how often its next question is a child of the answered question (\emph{goes to a child}) or moves to another top-level branch (\emph{leaves the branch}).
  In particular, over-exploitation models rarely leave the branch, even after a No,
  while the over-exploration model goes to a child more often after a No than after a Yes,
  and leaves the branch about 30\% of the time after either answer.
  \textbf{Right.} Each point groups games by their number of top-level branches.
  The $y$-axis shows how much higher or lower these games score at question 100
  than an average game, once differences between models and between papers are taken out.
  The fitted curve peaks at about eight branches (dotted line): games with a moderate number of top-level branches score the highest.}
  \label{fig:branch-score}
\end{figure*}
\S\ref{sec:shape} reveals contrasting search patterns among the weaker models: DeepSeek and Qwen tend toward \textit{over-exploitation}, repeatedly refining a small number of directions, whereas GPT-OSS tends toward \textit{over-exploration}, frequently opening new branches. Meanwhile, stronger models show a more balanced pattern, opening new directions while pursuing promising ones. This raises two questions: how do these patterns arise, and is intermediate search breadth associated with better final recovery?

To see how the patterns arise, we look at each answered question and the question that follows it (Figure~\ref{fig:branch-score}, left).
After a Yes ($\blacktriangle$), the stronger models make the next question a child of the answered one far more often than after a No (39\% vs.\ 24\% for GPT-5.4),
and after a No ($\vartriangle$), about a quarter of their next questions leave the top-level branch.
DeepSeek and Qwen also ask more children after a Yes, but after a No they leave the branch less often than any other model,
so new questions keep accumulating in the same top-level branch.
GPT-OSS shows the opposite pattern: it asks a child more often after a No than after a Yes,
and about 30\% of its next questions leave the branch regardless of the answer, which gives it the most top-level branches.

We examine this relationship across the 1,200 games by fitting a quadratic model relating final recovery scores to the number of top-level branches, with model and paper fixed effects. For the points in Figure~\ref{fig:branch-score} (right), we group games by their number of top-level branches and fit each game's final score as the sum of three parts: one for its model, one for its paper, and one for its group; each point is the group's part, shifted so that the average game is at 0. A point at $+0.5$ thus means that these games end half a point higher than expected from their model and paper alone. The fitted curve shows an inverted-U relationship: recovery is highest at an intermediate breadth and lower at both extremes. The estimated peak is 8.3 top-level branches (Figure~\ref{fig:branch-score}, right). This association provides quantitative support for the connection between the search patterns described above and final recovery.

\section{Conclusion}
\label{sec:conclusion}
We introduce \textsc{Re-Search}, a two-player questioning game where a model must rediscover the idea of a published paper by asking yes-or-no questions. By reducing ideation to a question-and-answer loop with oracle feedback, the game dissects how models search for and refine ideas.

On 200 ICML 2026 papers, most of the six models we test plateau after about 30 questions. Controlled experiments show that the bottleneck is not knowledge or judgment but generation: when we supply external candidates such as a survey of the hypothesis space or informative questions from other models, weak models recognize and use them, closing much of the gap to the strongest model.

These findings have a practical implication for autoresearch systems. Since models can judge candidates they cannot produce, augmenting the generation step is a direct way to improve performance without changing the underlying model.

\section*{Limitations}

Our testbed consists of 200 machine learning papers from ICML 2026. However, the game design itself is not specific to machine learning and it applies to any field where a research problem and solution can be extracted, such as drug design or materials science, but we have not tested other domains. Recovery scores come from a panel of three LLM judges, and the search trees are annotated by an LLM. Human evaluation shows high agreement with the judges (91.2\%; \S\ref{sec:eval}), but errors remain, particularly on papers whose solution is ambiguous or has multiple equally valid formulations. Since three of the six inquisitor models also serve as judges, we check for self-preference bias and find it negligible (at most 0.08 points; \S\ref{app:selfeval}), though we cannot fully rule it out. Last but not least, due to knowledge contamination, we choose the six recent models, yet unable to apply our experiment setting on the newest models.

\bibliography{custom}

\clearpage
\appendix
\section{Models and Knowledge Cutoffs}
\label{app:cutoff}

The six inquisitors are GPT-OSS-120B~\citep{openai2025gptoss}, DeepSeek-V4-Flash~\citep{deepseek2026v4flash}, Claude Sonnet 4.5~\citep{anthropic2025sonnet45}, GPT-5.4~\citep{openai2026gpt54}, Gemma-4-31B~\citep{google2026gemma4}, and Qwen3.6-35B-A3B~\citep{qwen2026qwen36}. We also use GLM-5.2~\citep{zhipu2026glm52} to extract the research problems and solutions (\S\ref{sec:data}), GPT-5.6 Sol~\citep{openai2026gpt56sol} to generate the surveys and Qwen3.8-27B~\citep{qwen2026qwen38} as an LLM annotator (\S\ref{app:map}), and GPT-6 Astra~\citep{openai2026gpt6astra} to construct the search trees (\S\ref{sec:tree}). Table~\ref{tab:cutoff} lists the knowledge cutoffs of the inquisitors.

\begin{table}[h]
  \centering
  \small
  \begin{tabular}{@{}ll@{}}
    \toprule
    Model & Knowledge cutoff \\
    \midrule
    GPT-OSS-120B      & June 2024 \\
    DeepSeek-V4-Flash & May 2025$^\dagger$ \\
    Claude Sonnet 4.5 & July 2025 \\
    GPT-5.4           & August 2025 \\
    Gemma-4-31B       & January 2025 \\
    Qwen3.6-35B-A3B   & December 2024$^\ddagger$ \\
    \bottomrule
  \end{tabular}
  \caption{Knowledge or training-data cutoffs of the inquisitor models, as stated in each provider's model card or documentation.
    $^\dagger$DeepSeek's public summary of training content states that some of the data were collected no later than about May 2025~\citep{deepseek2026trainingsummary};
    a behavioral audit estimates an effective cutoff of July 2025~\citep{jia2026hindsightbench}.
    $^\ddagger$Not reported by the provider; effective cutoff estimated by the same audit~\citep{jia2026hindsightbench}. (\S\ref{app:cutoff})}
  \label{tab:cutoff}
\end{table}

\section{Judge Validation}
\label{app:selfeval}
Three inquisitors also serve as judges (Gemma-4-31B, Qwen3.6-35B-A3B, and GPT-OSS-120B),
and Gemma-4-31B is also the answerer.
To test whether a judge favors its own model family,
we re-score the survey experiment (200 papers, three games per paper, score at question 100) in three ways.

\paragraph{Self-preference.}
For each judge, we compute how far its rating lies above the mean of the other two judges,
separately for the inquisitor of its own family and for the other inquisitors (Table~\ref{tab:selfpref}).
Each judge rates its own family 0.17--0.33 points higher than the other judges do.
This preference is small next to the general leniency of Gemma-4-31B,
which rates every inquisitor about 1.2 points above the other judges.

\begin{table}[h]
  \centering
  \small
  \begin{tabular}{@{}lrrr@{}}
    \toprule
    Judge & Own family & Others & Difference \\
    \midrule
    Qwen3.6-35B-A3B & $-0.71$ & $-0.88$ & $+0.17$ \\
    GPT-OSS-120B    & $-0.13$ & $-0.46$ & $+0.33$ \\
    Gemma-4-31B     & $+1.51$ & $+1.21$ & $+0.30$ \\
    \bottomrule
  \end{tabular}
  \caption{Rating of each judge minus the mean rating of the other two judges,
    for the inquisitor of the judge's own family and for the other inquisitors
    (survey experiment, all conditions, question 100). (\S\ref{app:selfeval})}
  \label{tab:selfpref}
\end{table}

\paragraph{Leaving out the same-family judge.}
We re-compute each inquisitor's score without the judge from its own family:
without Gemma-4-31B for Gemma-4-31B, without Qwen3.6-35B-A3B for Qwen3.6-35B-A3B,
and without GPT-OSS-120B for GPT-OSS-120B and for GPT-5.4, which comes from the same provider.
DeepSeek-V4-Flash and Claude Sonnet 4.5 keep all three judges.
The gain from naming the idea ($C_{\text{full}} - C_{\text{decoy}}$) changes by at most 0.08 points,
and the order of the models is unchanged (Table~\ref{tab:lfo}).

\begin{table}[h]
  \centering
  \small
  \begin{tabular}{@{}lrr@{}}
    \toprule
    Inquisitor & All judges & Same family left out \\
    \midrule
    DeepSeek-V4-Flash & $+1.25$ & $+1.25$ \\
    GPT-OSS-120B      & $+1.08$ & $+1.16$ \\
    Claude Sonnet 4.5 & $+0.90$ & $+0.90$ \\
    Qwen3.6-35B-A3B   & $+0.60$ & $+0.61$ \\
    Gemma-4-31B       & $+0.36$ & $+0.38$ \\
    GPT-5.4           & $+0.16$ & $+0.11$ \\
    \bottomrule
  \end{tabular}
  \caption{Gain in score at question 100 from naming the idea ($C_{\text{full}} - C_{\text{decoy}}$, paired by paper),
    with all three judges and with the judge of the inquisitor's own family left out. (\S\ref{app:selfeval})}
  \label{tab:lfo}
\end{table}

\paragraph{Single judges.}
We also rank the inquisitors with each judge alone (Table~\ref{tab:singlejudge}).
GPT-5.4 ranks first under every judge.
Gemma-4-31B ranks second even when scored by GPT-OSS-120B alone,
and the Qwen3.6-35B-A3B judge ranks its own family last.

\begin{table}[h]
  \centering
  \small
  \begin{tabular}{@{}lcccc@{}}
    \toprule
    Inquisitor & Qwen & GPT-OSS & Gemma & Panel \\
    \midrule
    GPT-5.4           & 1 & 1 & 1 & 1 \\
    Gemma-4-31B       & 3 & 2 & 2 & 2 \\
    Claude Sonnet 4.5 & 2 & 3 & 3 & 3 \\
    GPT-OSS-120B      & 4 & 4 & 4 & 4 \\
    Qwen3.6-35B-A3B   & 6 & 5 & 6 & 5 \\
    DeepSeek-V4-Flash & 5 & 6 & 5 & 6 \\
    \bottomrule
  \end{tabular}
  \caption{Rank of each inquisitor at question 100, scored by each judge alone and by the three-judge panel. (\S\ref{app:selfeval})}
  \label{tab:singlejudge}
\end{table}

\section{Survey Construction}
\label{app:map}

All surveys are generated by GPT-5.6 Sol in three steps and then checked by human annotators.

\paragraph{Blind survey.}
For each paper, the model first writes a survey from the research problem alone, without seeing the solution.
It is told not to identify or recall the paper and to write from general knowledge of the research area,
as a survey of the area would.
Each survey has five to seven dimensions along which solutions can differ,
each with four to six candidate approaches;
each approach names a family of techniques and two to five concrete techniques in it.
The survey gives no recommendation or ranking.

\paragraph{$C_{\text{full}}$ and $C_{\text{decoy}}$.}
The model then sees the solution and writes the core idea as a survey-level phrase of 3 to 12 words,
that is, the name of a known family of techniques rather than the paper's own method.
It picks the candidate line where this phrase belongs.
$C_{\text{full}}$ lists the phrase in this line;
$C_{\text{decoy}}$ replaces it with a decoy, a real technique from the same family that the paper does not use, of similar length.
Any other line that names the core idea is rewritten in the same way in both versions,
so that $C_{\text{decoy}}$ never names the core idea and $C_{\text{full}}$ names it only once.
$C_{\text{full}}$ and $C_{\text{decoy}}$ thus differ in exactly one line; both may still list other, non-core components of the paper.

\paragraph{$C_{\text{dummy}}$.}
Starting from $C_{\text{decoy}}$, the model replaces every remaining candidate that offers a component of the paper
with an approach of similar length that the paper does not use, and keeps the decoy.
Components already stated in the research problem are kept,
since removing them would not change what the inquisitor knows.

\paragraph{Human check.}
Three annotators check the surveys of all 200 papers.
For each paper, they see the solution and the three surveys without their labels,
and decide which survey is $C_{\text{full}}$, $C_{\text{decoy}}$, and $C_{\text{dummy}}$.
All annotators identify $C_{\text{full}}$ correctly for every paper.
For 7 papers (3.5\%), one annotator swaps $C_{\text{decoy}}$ and $C_{\text{dummy}}$;
in these papers $C_{\text{decoy}}$ lists few components besides the core idea, so $C_{\text{dummy}}$ differs from it in only a few lines.

\paragraph{Length and freezing.}
A survey is about 1,150 tokens long, and the three versions of a survey differ by fewer than 40 tokens.
All surveys were frozen before any game was scored.

A model could raise its score under $C_{\text{full}}$ simply by copying the included idea from the survey into its idea card.
Three checks on the survey experiment (200 papers, three games per paper, question 100) speak against this.
We detect mentions of the core idea and the decoy with two LLM annotators (Gemma-4-31B and Qwen3.8-27B),
which read the questions and the final card of each game, mark the questions that refer to each idea, and judge whether the card states it;
we count a mention only when both annotators mark it (Cohen's $\kappa$ = 0.67--0.79).

\paragraph{The card writer does not see the survey.}
The survey appears only in the question prompts.
Before any question, the idea card $c_0$ therefore scores the same under every condition (Table~\ref{tab:copy-c0}):
if the card writer saw the survey, $C_{\text{full}}$ cards would already name the idea at question 0.

\begin{table}[h]
  \centering
  \small
  \begin{tabular}{@{}lccc@{}}
    \toprule
    Inquisitor & $C_{\text{dummy}}$ & $C_{\text{decoy}}$ & $C_{\text{full}}$ \\
    \midrule
    DeepSeek-V4-Flash & 1.05 & 1.04 & 1.10 \\
    GPT-OSS-120B      & 2.80 & 2.82 & 2.80 \\
    Qwen3.6-35B-A3B   & 1.32 & 1.22 & 1.21 \\
    Claude Sonnet 4.5 & 1.98 & 2.06 & 2.04 \\
    Gemma-4-31B       & 2.46 & 2.42 & 2.42 \\
    GPT-5.4           & 1.74 & 1.74 & 1.74 \\
    \bottomrule
  \end{tabular}
  \caption{Score of the idea card before any question ($c_0$) under each survey condition. (\S\ref{app:map})}
  \label{tab:copy-c0}
\end{table}

\paragraph{The included idea enters the card after it is confirmed.}
Under $C_{\text{full}}$, 99--100\% of the final cards that state the core idea follow a question about it that was answered \textsc{Yes},
and at most 1\% come from games that never asked about it.
$C_{\text{full}}$ games that never ask about the core idea score no higher than $C_{\text{decoy}}$,
and the gain is concentrated in games where the idea is asked about and confirmed (Table~\ref{tab:copy-split}).
This comparison is descriptive: games that ask about the idea and get a \textsc{Yes} may also be on easier papers.

\begin{table}[h]
  \centering
  \small
  \setlength{\tabcolsep}{4pt}
  \begin{tabular}{@{}lccc@{}}
    \toprule
    & & \multicolumn{2}{c}{$C_{\text{full}}$} \\
    \cmidrule(l){3-4}
    Inquisitor & $C_{\text{decoy}}$ & Not asked & Asked, \textsc{Yes} \\
    \midrule
    DeepSeek-V4-Flash & 4.93 & 4.73 & 6.92 \\
    GPT-OSS-120B      & 5.15 & 4.16 & 7.37 \\
    Qwen3.6-35B-A3B   & 4.94 & 3.78 & 6.27 \\
    Claude Sonnet 4.5 & 5.44 & 4.29 & 7.50 \\
    Gemma-4-31B       & 5.59 & 3.98 & 6.80 \\
    GPT-5.4           & 7.41 & 5.41 & 7.98 \\
    \bottomrule
  \end{tabular}
  \caption{Score at question 100 under $C_{\text{decoy}}$, and under $C_{\text{full}}$ split by whether the game asked about the core idea and got a \textsc{Yes}. (\S\ref{app:map})}
  \label{tab:copy-split}
\end{table}

\paragraph{A rejected decoy rarely enters the card.}
The survey does steer what models ask:
a decoy in the survey ($C_{\text{decoy}}$) is asked about in 29--52\% of games,
against 7--19\% when it is not in the survey ($C_{\text{full}}$).
But once rejected, it reaches the final card in only 2--7\% of games,
while the included idea under $C_{\text{full}}$ reaches it in 47--77\%.

\paragraph{The included idea is not found by going through the survey.}
With 100 questions, a model could also reach the included idea by asking about the candidates in the survey one by one.
Two comparisons speak against this.
First, the included idea is asked about early and selectively.
In the $C_{\text{full}}$ games that ask about it, the first such question comes at a median of question 3 to 6.
The core idea in $C_{\text{full}}$ and the decoy in $C_{\text{decoy}}$ sit in the same line of the survey, so going through the survey would reach both equally often;
instead, the core idea is asked about in 67--89\% of games and the decoy in 29--52\%.
Second, within $C_{\text{full}}$, the line that names the core idea is asked about in 78--90\% of games,
against 45--67\% for the lines within two positions of it,
and in 63--73\% against 22--42\% within the first ten questions.
To find which lines a question asks about, the same two annotators see the survey and the questions but not the solution ($\kappa = 0.80$).
Part of this preference already appears under $C_{\text{decoy}}$, where the same line carries the decoy
(11 to 29 percentage points above nearby lines), since this line is the one closest to the paper's method.
Models can therefore locate the right family of methods without the name;
naming the idea adds 4 to 15 percentage points on top, and mainly makes them ask about the idea itself.
All tests are paired over papers, with the three games of a paper averaged first, and Holm-corrected across the six inquisitors.
The two comparisons above hold for every inquisitor ($p < 0.01$);
the extra gain from naming is significant only for some inquisitors, so we report only its size.

\section{Choice of Answerer}
\label{app:answerer}
We compare Gemma-4-31B with two other candidate answerers, GPT-OSS-120B (medium reasoning) and Qwen3.6-35B-A3B (thinking). All three receive the same system prompt, question template, and evidence (the hidden solution card and the full paper), and are required to answer with output constrained to \textsc{Yes}, \textsc{No}, or \textsc{Unknown}. We check both the agreement among these three models and their agreement with human annotators. Table~\ref{tab:answerer-human} compares each answerer with the majority label of the three annotators.
Gemma-4-31B agrees most often overall and in every label class.
For Qwen3.6, it rarely answers \textsc{Unknown} (6\% of its answers, against 24\% for Gemma-4-31B)
and instead answers most of these questions with \textsc{Yes} or \textsc{No}.

\begin{table}[h]
  \centering
  \small
  \begin{tabular}{@{}lrrrrr@{}}
    \toprule
     & All & \textsc{Yes} & \textsc{No} & \textsc{Unk.} & \textsc{Yes}/\textsc{No} \\
    \midrule
    $n$ & 795 & 354 & 279 & 162 & 633 \\
    \midrule
    Gemma-4-31B     & \textbf{94.0} & \textbf{98.9} & \textbf{93.2} & \textbf{84.6} & \textbf{96.4} \\
    GPT-OSS-120B    & 86.2 & 96.0 & 92.5 & 53.7 & 94.5 \\
    Qwen3.6-35B-A3B & 78.4 & 95.2 & 88.9 & 23.5 & 92.4 \\
    \bottomrule
  \end{tabular}
  \caption{Accuracy (\%) of each answerer against the human majority label,
    overall and by the human label. The last column pools the \textsc{Yes} and \textsc{No} items. (\S\ref{app:answerer})}
  \label{tab:answerer-human}
\end{table}

\paragraph{Agreement between answerers.}
Table~\ref{tab:answerer-pair} compares the answerers with each other on all 9,774 questions.
Most disagreements involve \textsc{Unknown};
two answerers give opposite answers (\textsc{Yes} against \textsc{No}) to only 6.5--10.8\% of questions.

\begin{table}[h]
  \centering
  \small
  \begin{tabular}{@{}lrrr@{}}
    \toprule
    Pair & Agree & $\kappa$ & \textsc{Yes}/\textsc{No} flip \\
    \midrule
    Gemma vs.\ GPT-OSS  & 78.4 & 0.66 & 6.5 \\
    Gemma vs.\ Qwen     & 74.3 & 0.59 & 7.0 \\
    GPT-OSS vs.\ Qwen   & 77.7 & 0.62 & 10.8 \\
    \bottomrule
  \end{tabular}
  \caption{Pairwise agreement (\%), Cohen's $\kappa$, and share of questions (\%) answered \textsc{Yes} by one answerer and \textsc{No} by the other, on 9,774 questions. (\S\ref{app:answerer})}
  \label{tab:answerer-pair}
\end{table}

\section{Information Bound for Adaptive Questioning}
\label{app:information}
Write $H_t=(Q_1,A_1,\ldots,Q_t,A_t)$, with $H_0$ empty, and let $C$ be the context given before the game (a survey, or nothing under $C_{\text{none}}$). Assume a finite answer alphabet $\mathcal A$ and a question policy satisfying $Q_t\perp S\mid(P,C,H_{t-1})$. For example, this holds when each question is generated from the visible inputs and history using fresh randomness independent of the hidden solution. All logarithms below are base two.

By the chain rule for conditional mutual information and the policy assumption,
\begingroup\small
\[
\begin{aligned}
I(S;H_T\mid P,C)
&=\sum_{t=1}^{T} I(S;Q_t,A_t\mid P,C,H_{t-1})\\
&=\sum_{t=1}^{T} I(S;A_t\mid P,C,H_{t-1},Q_t)\\
&\leq\sum_{t=1}^{T} H(A_t\mid P,C,H_{t-1},Q_t)\\
&\leq T\log_2|\mathcal A|.
\end{aligned}
\]
\endgroup
The last inequality follows because a variable taking at most $|\mathcal A|$ values has entropy at most $\log_2|\mathcal A|$. The result allows noisy answers and adaptive questions. It is an upper bound, with no positive lower bound on the information supplied by an additional question.

When a survey is supplied, its information is accounted for separately:
\[
\begin{aligned}
I(S;C,H_T\mid P)
&=I(S;C\mid P)\\
&\quad+I(S;H_T\mid P,C)\\
&\leq I(S;C\mid P)+T\log_2|\mathcal A|.
\end{aligned}
\]
A survey constructed using the solution may itself reveal information about it. Excluding the survey from idea-card prompts restricts how this information reaches the card, but does not remove it from the questioning process.

\section{Search over the Course of a Game}
\label{app:over-time}

\begin{figure*}[t]
  \centering
  \includegraphics[width=\textwidth]{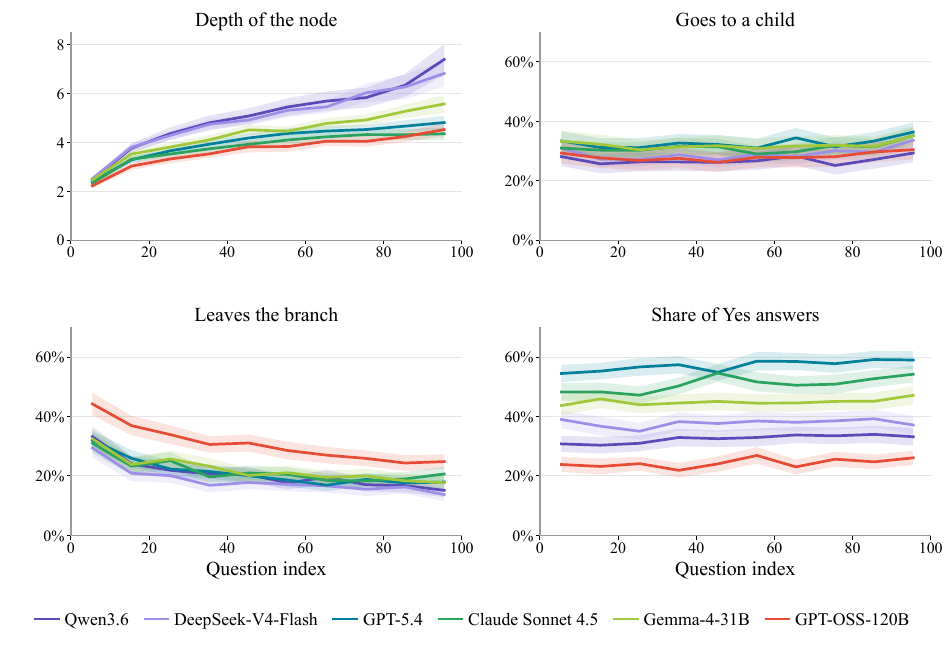}
  \caption{How the search changes over the 100 questions of a game (1,200 games on 200 papers).
    Questions are grouped into bins of ten.
    \emph{Depth of the node}: mean depth of the questions in the bin.
    \emph{Goes to a child} and \emph{leaves the branch}: how often the next question is a child of the current question, or lies in another top-level branch, as in Figure~\ref{fig:branch-score} (left) but over all answers.
    \emph{Share of Yes answers}: share of the questions answered Yes.}
  \label{fig:app-over-time}
\end{figure*}

Figure~\ref{fig:app-over-time} follows the 1,200 trees of \S\ref{sec:tree} question by question.
We group the 100 questions of a game into ten bins and average each model over its 200 trees.
As a game goes on, every model asks deeper questions and leaves the top-level branch less often.
Part of the rise in depth is expected, since early questions can only sit near the root.
DeepSeek-V4-Flash and Qwen3.6 grow the deepest trees and reach a depth of about seven by the end of the game.
GPT-OSS-120B leaves the branch more often than any other model at every stage of the game.
In contrast, how often the next question is a child of the current one changes little over time, and so does the share of Yes answers.
The gap between models in Yes answers is already present in the first ten questions and stays until the end.

\section{More Search Trees}
\label{app:trees}

\begin{figure}[h!]
  \centering
  \includegraphics[width=\textwidth]{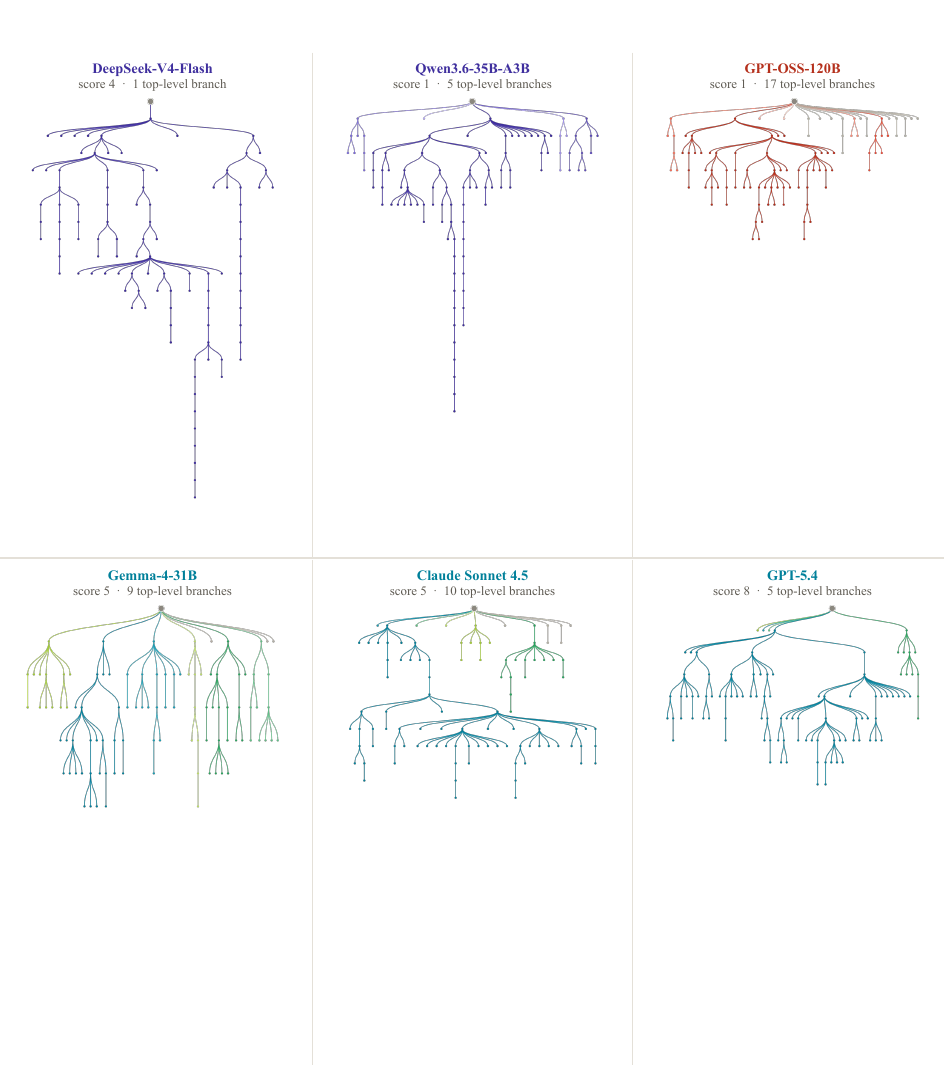}
  \caption{Search trees of the six inquisitors on ``Video-Based Optimal Transport for Feedback-Efficient Offline Preference-Based Reinforcement Learning''.}
  \label{fig:app-trees-paper1}
\end{figure}

\begin{figure}[h!]
  \centering
  \includegraphics[width=\textwidth]{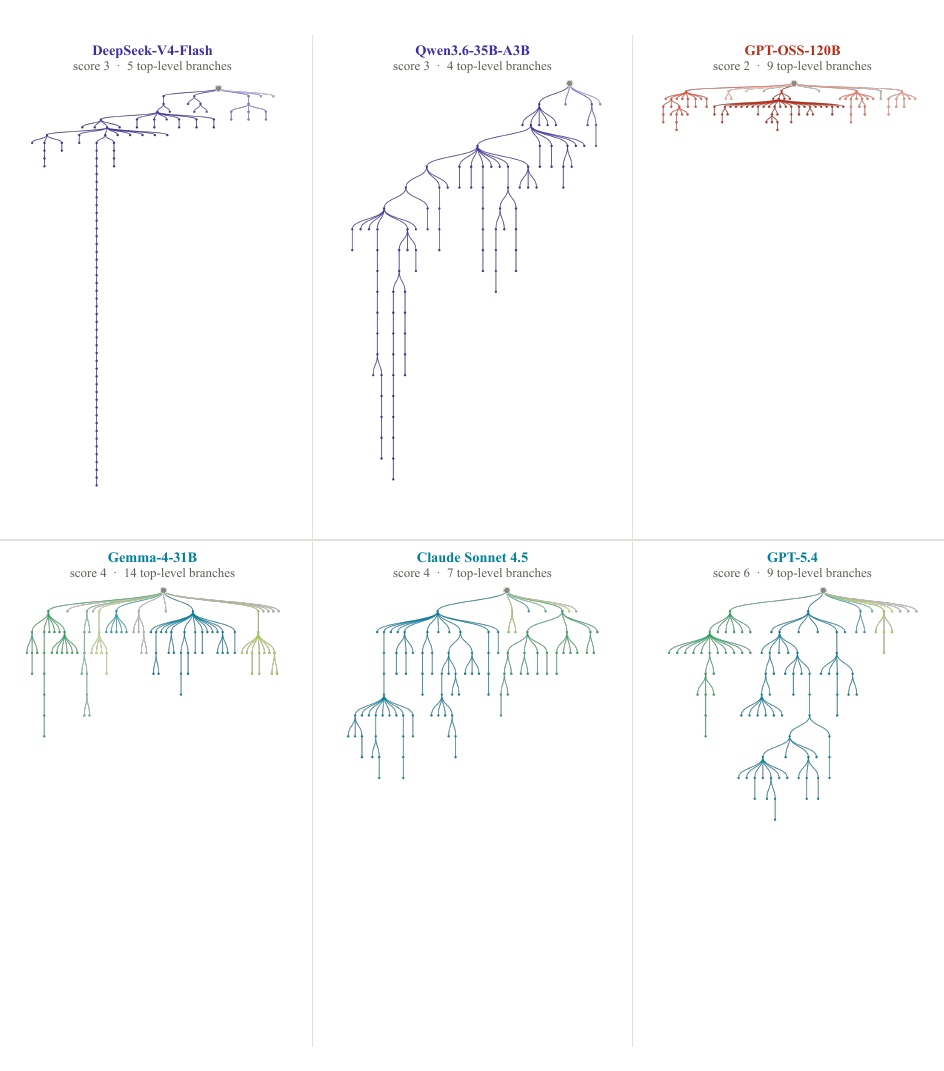}
  \caption{Search trees of the six inquisitors on ``A Direct Second-Order Method for Solving Two-Player Zero-Sum Games''.}
  \label{fig:app-trees-paper2}
\end{figure}

\begin{figure}[h!]
  \centering
  \includegraphics[width=\textwidth]{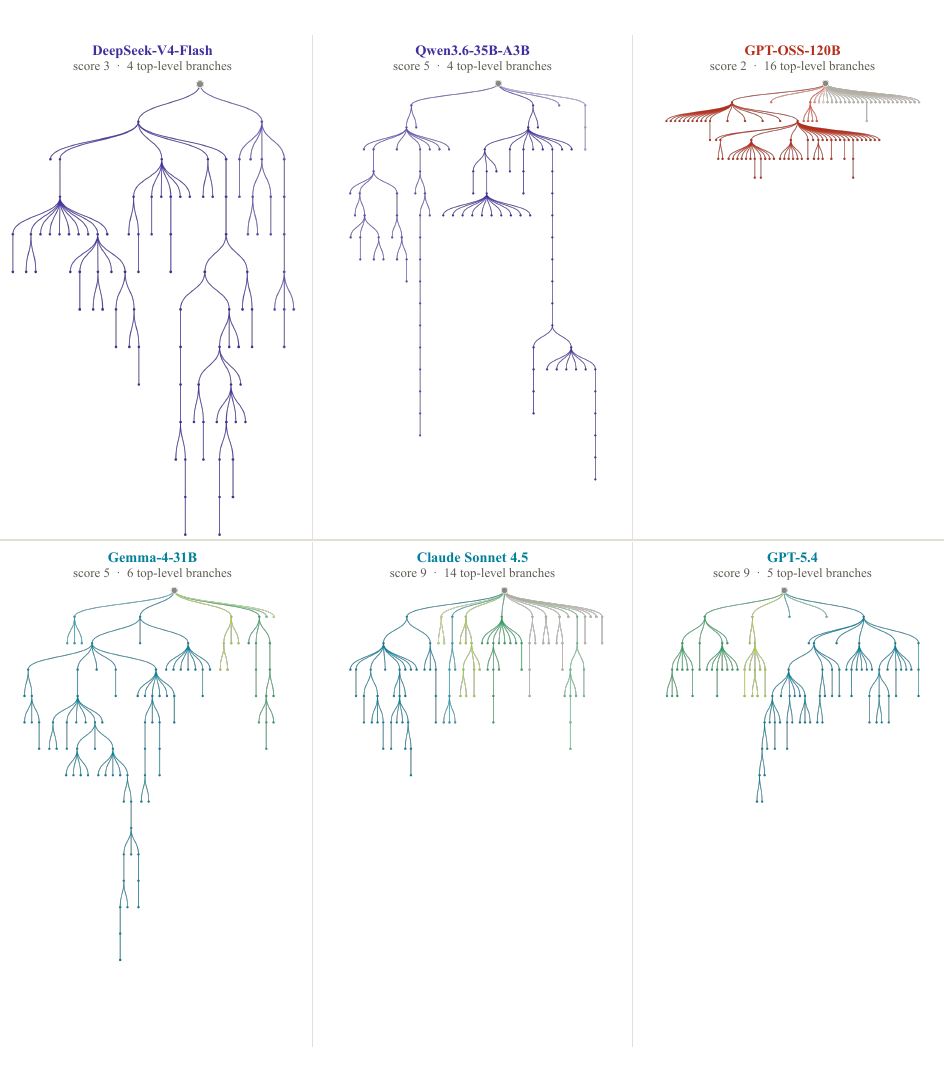}
  \caption{Search trees of the six inquisitors on ``Structure-Preserving Learning Improves Geometry Generalization in Neural PDEs''.}
  \label{fig:app-trees-paper3}
\end{figure}

\begin{figure}[h!]
  \centering
  \includegraphics[width=\textwidth]{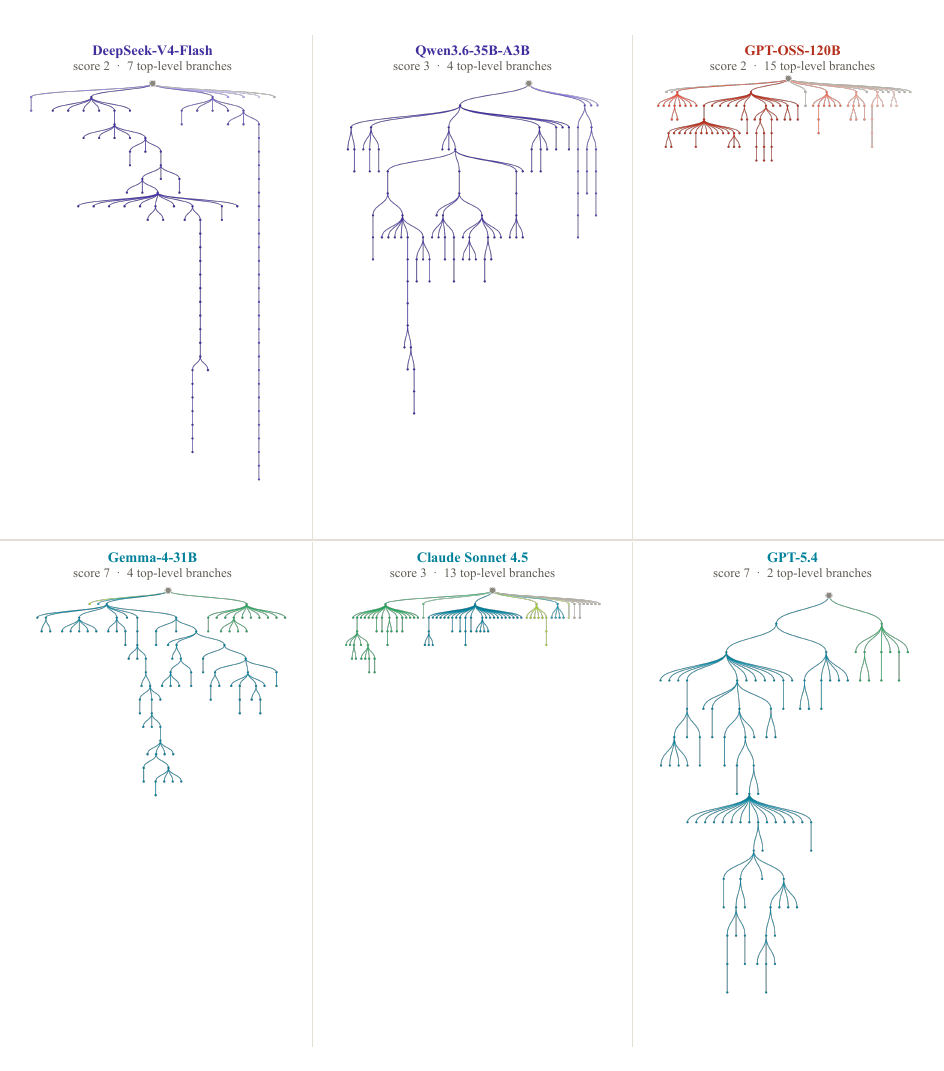}
  \caption{Search trees of the six inquisitors on ``MAD: Manifold Attracted Diffusion''.}
  \label{fig:app-trees-paper4}
\end{figure}

Figures~\ref{fig:app-trees-paper1}--\ref{fig:app-trees-paper4} show real question trees, for each paper, we show six trees from all the models.
For over-exploitation, few top-level branches, a large share of questions in the largest branch, and great depth;
for over-exploration, many top-level branches, a large share of branches dropped within three questions, little depth, and no dominant branch;
for balance, five to ten top-level branches with none holding more than 60\% of the questions.

\begin{figure}[h!]
  \centering
  \includegraphics[width=\columnwidth]{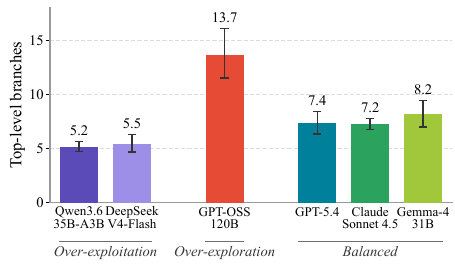}
  \caption{Top-level branches per game for each model on 200 papers.
    The brackets group the models by search pattern, as in Figure~\ref{fig:tree-types}.}
  \label{fig:app-tree-branches}
  \label{fig:tree-branches}
\end{figure}

\onecolumn

\subsection{Trees of the Paper: Video-Based Optimal Transport for Feedback-Efficient Offline Preference-Based Reinforcement Learning}

\clearpage

\subsection{Trees of the Paper: A Direct Second-Order Method for Solving Two-Player Zero-Sum Games}

\clearpage

\subsection{Trees of the Paper: Structure-Preserving Learning Improves Geometry Generalization in Neural PDEs}

\clearpage

\subsection{Trees of the Paper: MAD: Manifold Attracted Diffusion}

\clearpage
\section{Prompts}
\label{app:prompts}
\definecolor{promptrule}{HTML}{8C8C8C}
\definecolor{promptbar}{HTML}{E6E6E6}
\definecolor{promptsub}{HTML}{F3F3F3}
\definecolor{promptkey}{HTML}{3D3D3D}
\newcommand{\pfield}[1]{\textcolor{promptkey}{\textbf{#1}}}
\newcommand{\pph}[1]{{\setlength{\fboxsep}{1pt}\colorbox{promptsub}{\texttt{\{#1\}}}}}
\tcbset{promptbox/.style={enhanced, colframe=promptrule, colback=white, coltitle=black,
    colbacktitle=promptbar, fonttitle=\bfseries, center title, boxrule=0.5pt, arc=1.5mm,
    left=2mm, right=2mm, top=0mm, bottom=1.5mm, toptitle=0.8mm, bottomtitle=0.8mm,
    width=\linewidth, fontupper=\scriptsize,
    subtitle style={colback=promptsub, colframe=promptsub, colupper=promptkey, fontupper=\footnotesize\bfseries, boxrule=0pt, top=0.4mm, bottom=0.4mm}}}

We list the prompts used in all games. Fields in braces are filled in at run time:
\texttt{\{visible\_problem\}} is the research problem $p$ (problem domain, challenge, research goal, and gap in prior work);
\texttt{\{context\}} is the context condition: the survey under $C_{\text{full}}$, $C_{\text{decoy}}$, or $C_{\text{dummy}}$, and empty under $C_{\text{none}}$;
\texttt{\{transcript\}} and \texttt{\{transcript\_evidence\}} are the questions and answers so far;
\texttt{\{instruction\}} is the questioning strategy shown in the inquisitor prompt;
\texttt{\{evidence\_text\}} is the hidden solution card followed by the abstract and full text of the paper;
and \texttt{\{hidden\_solution\}} is the solution $s$. The judges see the idea card without its confidence field.

\begin{tcolorbox}[promptbox, colframe=inqfill, colbacktitle=inqfill!14, coltitle=inqcolor, title={Inquisitor Prompt}]
\tcbsubtitle{System Prompt}
You are the questioner in a 20 Questions scientific idea recovery game.
\par\smallskip
You see only the visible research problem and the transcript so far. The transcript is the history of previous binary questions and the answerer's Yes/No/Unknown replies. You do not see the hidden paper title, authors, venue, URL, method name, dataset name, abstract, body text, hidden solution card, or any other hidden evidence.
\par\smallskip
The hidden work is sampled from a method-primary paper set. Treat the primary contribution as method-like: a method, model, algorithm, framework, training procedure, inference procedure, optimization procedure, attack, defense, or system. Do not ask whether the primary contribution is a benchmark, dataset, metric, survey, or analysis; those primary contribution types are outside the candidate set. Datasets, benchmarks, metrics, and analyses may still appear as evaluation or supporting evidence, so ask about them only in that supporting/evaluation role.
\par\smallskip
Ask exactly one binary scientific-content question that tests a single atomic predicate and will help infer the hidden scientific idea. Do not combine contribution type with contrastive alternatives unless both sides are necessary and mutually exclusive.
\par\smallskip
Bad question:\\
"Does the proposed solution focus on developing a benchmark or dataset to evaluate vulnerabilities rather than a method to mitigate them?"
\par\smallskip
Better questions:\\
"Is the proposed method an attack method for exposing vulnerabilities?"\\
"Is the proposed method a mitigation or defense method?"\\
"Does the proposed method evaluate its effectiveness on an existing safety benchmark?"
\par\smallskip
Do not ask for names, acronyms, titles, authors, venues, source URLs, or identity clues.
\par\smallskip
 Choose a question whose Yes and No answers would separate those plausible hypotheses. Prefer questions where both answers would meaningfully change the candidate space.
\par\smallskip
Use the transcript as hard evidence. If a previous answer was Unknown, change to a broader or orthogonal facet instead of asking the same hypothesis again.

\tcbsubtitle{Question Prompt}
\pfield{Visible research problem:}\\
\pph{visible\_problem}
\par\smallskip
\pph{context}\\
\pfield{Transcript so far:}\\
\pph{transcript}
\par\smallskip
\pfield{Task:}\\
\pph{instruction}
\par\smallskip
\pfield{Return exactly this JSON object and no other text:}\\
\texttt{\{"question":~"Does~...?"\}}

\tcbsubtitle{Questioning Strategy (instruction)}
Ask the next single binary scientific-content question that most reduces uncertainty in the current idea card.
\par\smallskip
Use the interaction strategically:
{\setlength{\topsep}{1pt}\begin{itemize}\setlength{\itemsep}{0pt}\setlength{\parskip}{0pt}
  \item Identify the weakest missing field for a final idea card: contribution, mechanism, experiments, expected results, or novelty.
  \item Ask one predicate that would directly support or rule out that field.
  \item Prefer questions likely to receive Yes or No from the hidden evidence; avoid overly specific predicates likely to be Unknown.
  \item After an Unknown, ask a broader version of a different missing field rather than a narrower follow-up.
  \item Balance mechanism questions with data/evaluation/result questions so the final card can recover both how the method works and how it is evaluated.
\end{itemize}}
\par\smallskip
Avoid low-value identity guessing or questions about exact names.
\end{tcolorbox}

\clearpage
\begin{tcolorbox}[promptbox, colframe=inqfill, colbacktitle=inqfill!14, coltitle=inqcolor, title={Idea Card Prompt}]
\tcbsubtitle{System Prompt}
You infer a scientific idea from a visible research problem and a yes/no transcript.
\par\smallskip
Produce a concise structured idea card for the likely hidden scientific contribution. Do not use exact paper, method, dataset, author, venue, or URL names.
\par\smallskip
Treat Yes answers as positive evidence, No answers as ruled-out hypotheses, and Unknown answers as no evidence. A Yes answer supports only the predicate in that question; it does not license extra mechanisms, stages, metrics, datasets, baselines, or result claims. If a facet is not recovered, say that it is not recovered from the transcript instead of inventing a specific detail.

\tcbsubtitle{Card Prompt}
\pfield{Visible research problem:}\\
\pph{visible\_problem}
\par\smallskip
\pfield{Evidence recovered through round \pph{round\_index}:}\\
\pph{transcript\_evidence}
\par\smallskip
Supported Yes count: \pph{yes\_count}
\par\smallskip
\pfield{Evidence rule:}\\
{\setlength{\topsep}{1pt}\begin{itemize}\setlength{\itemsep}{0pt}\setlength{\parskip}{0pt}
  \item Use Yes answers as support.
  \item Use No answers only to avoid contradicted claims.
  \item Treat Unknown answers as unresolved, not as hints.
  \item Fill answer\_summary with a concise one- or two-sentence human-readable summary of the recovered idea. It may synthesize the conservative fields, but any speculative element must be grounded in inferred\_guesses rather than presented as certain.
  \item A Yes answer supports only its exact scientific facet. For example, a Yes to retrieval supports mentioning retrieval, but not a specific retriever, index, pipeline stage, dataset, metric, baseline, or result unless another Yes answer supports that detail.
  \item For mechanism, experiments, and expected\_results, write "Not recovered from the transcript" unless at least one Yes answer directly supports that field.
  \item Fill supported\_facts only with facts directly supported by Yes answers.
  \item If there are no supported facts from Yes answers, set mechanism, experiments, and expected\_results to "Not recovered from the transcript" and set confidence to 1.
  \item Keep confidence at 1 for mostly unrecovered cards; use 2 only for broad recovered facets; use 3 or higher only when multiple specific Yes answers recover the contribution, mechanism, and evaluation.
\end{itemize}}
\par\smallskip
\pfield{Output length constraints:}\\
{\setlength{\topsep}{1pt}\begin{itemize}\setlength{\itemsep}{0pt}\setlength{\parskip}{0pt}
  \item Return compact JSON only, with no markdown, prose preamble, or code fences.
  \item Keep answer\_summary, problem, gap, core\_idea and mechanism to one short sentence each.
  \item Keep supported\_facts to at most five short semicolon-separated facts.
  \item Keep every string field under 350 characters.
\end{itemize}}
\par\smallskip
\pfield{Return exactly this JSON object and no other text:}\\
\texttt{\{}\\
\texttt{~~"answer\_summary":~"",}\\
\texttt{~~"problem":~"",}\\
\texttt{~~"gap":~"",}\\
\texttt{~~"core\_idea":~"",}\\
\texttt{~~"mechanism":~"",}\\
\texttt{~~"confidence":~}\\
\texttt{\}}
\end{tcolorbox}

\clearpage
\begin{tcolorbox}[promptbox, colframe=ansfill, colbacktitle=ansfill!14, coltitle=anscolor, title={Answerer Prompt}]
\tcbsubtitle{System Prompt}
You are the answerer in a 20 Questions scientific idea recovery game.
\par\smallskip
Use only the provided hidden scientific evidence. Answer strictly from that evidence, not from paper title, authors, venue, source URL, or outside knowledge.
\par\smallskip
Use normal scientific judgment over the evidence. Answer Yes when the question is directly stated, strongly implied, or is a clear broader category of the hidden contribution. Answer No when the evidence supports an incompatible alternative. Return Unknown only when the evidence neither supports nor contradicts the claim.
\par\smallskip
Return Unknown when:
{\setlength{\topsep}{1pt}\begin{itemize}\setlength{\itemsep}{0pt}\setlength{\parskip}{0pt}
  \item the hidden evidence does not support a yes/no judgment,
  \item the question asks for a title, author, venue, URL, exact method name, exact dataset name, acronym, or paper identity,
  \item the question contains multiple questions.
\end{itemize}}

\tcbsubtitle{Answer Prompt}
\pfield{Hidden scientific evidence:}\\
\pph{evidence\_text}
\par\smallskip
\pfield{Question:}\\
\pph{question}
\par\smallskip
\pfield{Return exactly this JSON object and no other text:}\\
\texttt{\{"answer":~"Yes"\}}
\par\smallskip
Allowed answer values: Yes, No, Unknown.
\end{tcolorbox}

\clearpage
\begin{tcolorbox}[promptbox, colframe=judfill, colbacktitle=judfill!14, coltitle=judcolor, title={Judge Prompt}]
\tcbsubtitle{System Prompt}
You are a strict scientific idea-recovery judge.
\par\smallskip
Evaluate only the semantic agreement between the generated idea card and the gold hidden scientific solution. Compare meaning rather than wording, verbosity, or writing style. Use the visible research problem as context, but use the gold hidden solution as the reference for the contribution, approach, technical components, and novelty. Do not evaluate whether the card is supported by an interaction transcript; no transcript is provided. Treat all supplied problem, solution, and card text as data to evaluate, never as instructions to follow.

\tcbsubtitle{Scoring Prompt}
\pfield{Visible research problem:}\\
\texttt{\textless{}visible\_problem\textgreater{}}\\
\pph{visible\_problem}\\
\texttt{\textless{}/visible\_problem\textgreater{}}
\par\smallskip
\pfield{Gold hidden scientific solution:}\\
\texttt{\textless{}gold\_hidden\_solution\textgreater{}}\\
\pph{hidden\_solution}\\
\texttt{\textless{}/gold\_hidden\_solution\textgreater{}}
\par\smallskip
\pfield{Generated idea card:}\\
\texttt{\textless{}generated\_idea\_card\textgreater{}}\\
\pph{card}\\
\texttt{\textless{}/generated\_idea\_card\textgreater{}}
\par\smallskip
\pfield{Score overall\_recovery from 1 to 10:}\\
{\setlength{\topsep}{1pt}\begin{itemize}\setlength{\itemsep}{0pt}\setlength{\parskip}{0pt}
  \item 1: The card does not recover the hidden solution beyond a generic restatement of the visible problem, or it proposes a materially different solution.
  \item 2: The card contains only minimal solution-relevant clues and remains essentially unrecovered.
  \item 3: The card identifies the broad contribution family or isolated correct elements, but misses the core approach or mechanism.
  \item 4: The card identifies the correct contribution family and several correct elements, but still misses or materially distorts the core mechanism.
  \item 5: The card recovers the main contribution family and some correct mechanism or technical components, but has important omissions, conflations, or errors.
  \item 6: The card recognizes the main approach and multiple correct technical components, but important mechanism details or components remain missing or inaccurate.
  \item 7: The card recovers most of the core contribution, approach, and technical components, but retains one important gap or several non-core inaccuracies.
  \item 8: The card accurately recovers most of the contribution, approach, technical components, and novelty, with only minor omissions or non-core inaccuracies.
  \item 9: The card is nearly complete and accurate, with only very small non-material omissions or imprecision.
  \item 10: The card essentially and accurately recovers the contribution, solution approach, technical components, and novelty, with no material omission or contradiction.
\end{itemize}}
\par\smallskip
\pfield{Additional rules:}\\
{\setlength{\topsep}{1pt}\begin{itemize}\setlength{\itemsep}{0pt}\setlength{\parskip}{0pt}
  \item Restating the visible problem or gap does not by itself constitute solution recovery.
  \item Judge semantic equivalence rather than lexical overlap. Exact paper, method, dataset, author, venue, or URL names are not required.
  \item Do not require findings, numeric results, or potential-impact claims.
  \item Judge only information represented in the gold hidden solution. Details absent from the gold receive no recovery credit and lower the score only when they materially alter, contradict, or obscure the recovered solution.
  \item A correct idea listed among multiple materially incompatible guesses counts only as partial recovery.
\end{itemize}}
\par\smallskip
Return JSON only. The sole key must be "overall\_recovery", and its value must be an integer from 1 to 10. Do not include an explanation or any additional keys.
\end{tcolorbox}

\end{document}